\pdfoutput=1

\documentclass[11pt]{article}

\usepackage[]{acl}

\usepackage{times}
\usepackage{latexsym}

\usepackage[T1]{fontenc}
\usepackage[utf8]{inputenc}

\usepackage{microtype}

\usepackage{graphicx}
\usepackage{subcaption}
\usepackage{booktabs}
\usepackage{cleveref}
\usepackage{multirow}

\usepackage{todonotes}

\clearpage{}%
\newcommand{\insertGlueSpec}{
\begin{table}[!htb]
    \centering
    \resizebox{\columnwidth}{!}{%
    \begin{tabular}{lrl|lrl}
        \textbf{Dataset} & \textbf{|Train|} & \textbf{Metric} & \textbf{Dataset} & \textbf{|Train|} & \textbf{Metric} \\ \hline 
        MNLI & 393k & acc. & STS-B & 7k & Pearson/Spearman \\
        QQP & 364k & acc./F1 & MRPC & 3.7k & acc./F1 \\
        QNLI & 105k & acc. & RTE & 2.5k & acc.\\
        SST-2 & 67k & acc. & WNLI & 634 & acc. \\
         CoLA& 8.5k & Matthews & \\

    \end{tabular}
    }
    \caption{GLUE datasets with their number of training examples and the corresponding evaluation metric.}
    \label{tab:glue_data}
\end{table}
}

\newcommand{\insertArchitectureSelectionCrossDataRegime}{
\begin{table*}[!ht]
    \centering
     \resizebox{0.98\textwidth}{!}{%
    \begin{tabular}{lllllllllll}
     & \textbf{MNLI} &	\textbf{QQP} &	\textbf{QNLI} &	\textbf{SST-2} &	\textbf{CoLA} &  \textbf{STS-B} & \textbf{MRPC} & \textbf{RTE} & \textbf{WNLI}  & \textbf{Avg}\\ \hline

    \multicolumn{11}{c}{\textbf{Low-data-100}} \\ \hline
      \textbf{Baseline} & 33.89$_{3.02}$ & 30.65$_{0.38}$ & 58.78$_{4.81}$ &  56.01$_{3.68}$ & 5.20$_{4.84}$ & 40.00$_{9.64}$ & 74.80$_{0.0}$ & 49.39$_{2.86}$ & 55.21$_{3.01}$ & 44.87 \\
      \textbf{AA} & 33.64$_{2.66}$ & 30.88$_{0.39}$ & 59.61$_{6.19}$ & 51.28$_{2.52}$ & -0.55$_{1.87}$ & 45.18$_{14.17}$ & 74.80$_{0.0}$ & 50.11$_{3.44}$ &  55.48$_{2.46}$ & 44.49 \\
      \textbf{AdapterDrop$^{AA}$} & 33.72$_{2.84}$ & 30.62$_{0.40}$ & 57.50$_{5.78}$ & 54.01$_{2.59}$ & 4.10$_{7.95}$ & 36.53$_{8.93}$ & 74.8$_{0.0}$ & 49.39$_{2.86}$ & \bfseries 56.06$_{1.38}$ & 44.08 \\
      \textbf{AdapterDrop$^{13}$} & 33.71$_{2.76}$ & 30.61$_{0.4}$ & 58.39$_{4.27}$ & 53.44$_{2.56}$ & 3.91$_{7.6}$ & 36.23$_{8.68}$ & 74.8$_{0.0}$ & 49.46$_{2.81}$ & 55.76$_{1.91}$ & 44.04 \\

      \textbf{AA-focused$^{{spec}}$} &  35.28$_{2.06}$ &  44.37$_{16.31}$ & \bfseries 63.75$_{4.39}$ & 52.94$_{4.64}$ & 5.68$_{10.91}$ & \bfseries 62.79$_{3.34}$ & 74.80$_{0.01}$ & 51.48$_{2.72}$ & 54.08$_{4.51}$ &  49.47 \\
      \textbf{AA-focused$^{uni}$} & \bfseries 36.36$_{2.61}$ &  44.37$_{16.31}$ & 63.36$_{4.86}$ & 55.87$_{4.42}$ & 4.75$_{4.9}$ & 59.37$_{6.78}$ & \bfseries 74.94$_{0.2}$ & 51.12$_{3.45}$ & 51.83$_{4.12}$ & 49.11\\
      \textbf{AA-focused$^{sim}$} & 34.77$_{3.18}$ & \bfseries 45.78$_{14.40}$ & 63.13$_{4.30}$ & \bfseries 61.58$_{10.95}$ & \bfseries 17.54$_{11.19}$ & 59.89$_{7.70}$ & 74.77$_{0.07}$ & \bfseries 52.20$_{2.93}$ & 51.83$_{5.52}$ & \bfseries 51.28 \\
     \hline 
    \textbf{|Baseline|} & 24 & 24 & 24 & 24 & 24 & 24 & 24 & 24 & 24 & \\
    \textbf{|AA|} & 13.2$_{1.7}$ & 15.0$_{3.0}$ & 13.6$_{2.2}$ & 14.6$_{4.0}$ & 15.8$_{2.1}$ & 16.4$_{2.7}$ & 13.0$_{1.8}$ & 11.2$_{1.8}$ & 12.3$_{5.7}$ &\\ 
    \textbf{|AdapterDrop$^{AA}$|} & 14 & 13 & 15 & 16 & 16 & 14 & 15 & 13 & 16 \\
    \textbf{|AdapterDrop$^{13}$|} & 13 & 13 & 13 & 13 & 13 & 13 & 13 & 13 & 13 \\
    \textbf{|AA-focused$^{{spec}}$|} & 14 & 13 & 15 & 16 & 16 & 14 & 15 & 13 & 16 \\
    \textbf{|AA-focused$^{{uni}}$|} & 13 & 13 & 13 & 13 & 13 & 13 & 13 & 13 & 13 \\
    \textbf{|AA-focused$^{{sim}}$|} & 13 & 13 & 13 & 13 & 13 & 13 & 13 & 13 & 13 \\

      \hline
    \multicolumn{11}{c}{\textbf{Low-data-300}} \\ \hline
     \textbf{Baseline} & 36.55$_{4.76}$ & 61.50$_{8.66}$ & 69.62$_{1.24}$ & 79.86$_{14.15}$ & 30.40$_{5.48}$ & 78.24$_{2.81}$ & 76.55$_{1.31}$ & 51.62$_{3.21}$ & 45.92$_{4.33}$ & 58.91 \\
    \textbf{AA} & 37.14$_{2.49}$ & 66.07$_{0.38}$ & 71.33$_{1.82}$ & 72.52$_{16.59}$ & 26.05$_{8.74}$ & \bfseries 82.08$_{1.6}$ & 74.03$_{2.21}$ & 51.83$_{2.84}$ &  47.04$_{4.76}$ & 58.68 \\

      \textbf{AdapterDrop$^{AA}$} & 38.86$_{5.93}$ & 62.98$_{4.85}$ & 66.71$_{2.91}$ & 79.29$_{14.17}$ & 16.89$_{12.06}$ & 78.5$_{1.99}$ & 75.74$_{0.67}$ & 51.19$_{3.35}$ & 46.76$_{4.02}$ & 57.44\\
    \textbf{AdapterDrop$^{13}$} & 37.95$_{5.56}$ & 63.72$_{4.84}$ & 66.71$_{2.91}$ & 80.0$_{14.47}$ & 16.3$_{12.05}$ & 77.52$_{2.08}$ & 76.03$_{0.92}$ & 51.33$_{3.4}$ & 46.48$_{4.27}$ & 57.34 \\
     \textbf{AA-focused$^{{spec}}$} &  44.62$_{4.11}$ &  66.83$_{1.06}$ &  73.72$_{1.09}$ &  85.87$_{2.94}$ &  34.51$_{8.3}$ &  81.16$_{2.04}$ &  76.72$_{1.06}$ &  54.58$_{4.72}$ &  46.20$_{3.92}$ &  62.69 \\
      \textbf{AA-focused$^{uni}$} &  \bfseries 46.69$_{4.29}$ & \bfseries 69.25$_{1.33}$ & \bfseries 74.16$_{2.95}$ & \bfseries 87.57$_{0.72}$ &  35.65$_{3.26}$ &  81.71$_{2.64}$ & 75.97$_{1.55}$ & \bfseries 56.89$_{5.56}$ & \bfseries 52.39$_{7.26}$ & \bfseries 64.48 \\
      \textbf{AA-focused$^{sim}$} & 45.97$_{2.08}$ & 68.36$_{1.36}$ & 73.98$_{2.68}$ & 86.83$_{1.90}$ & \bfseries 37.43$_{3.10}$ & 78.81$_{3.58}$ & \bfseries 76.66$_{1.30}$ & 55.96$_{2.81}$ & 48.44$_{5.53}$ & 63.61 \\

     \hline
     \textbf{|AA|} & 17.0$_{1.3}$ & 16.2$_{1.0}$ & 14.8$_{1.8}$ & 12.8$_{3.2}$ & 16.8$_{2.2}$ & 18.6$_{1.9}$ & 16.0$_{1.1}$ & 12.4$_{1.2}$ & 12.4$_{2.1}$  \\
     \textbf{|AA-focused$^{spec}$|} & 18 & 16 & 13 & 9 & 17 & 16 & 16 & 13 & 13 \\

     \hline
     
    \multicolumn{11}{c}{\textbf{Low-data-500}} \\ \hline
     \textbf{Baseline} & 44.35$_{6.08}$ & 69.49$_{1.12}$ & 73.48$_{1.89}$ & 88.26$_{1.53}$ & 37.98$_{4.42}$ & 82.07$_{0.99}$ & 78.33$_{1.11}$ &  59.28$_{1.76}$ & 49.86$_{6.08}$ & 64.79\\
    \textbf{AA} &  47.33$_{5.11}$ & 67.52$_{2.99}$ & 75.0$_{2.3}$ & 84.93$_{3.06}$ & 39.96$_{4.87}$ & \bfseries 84.56$_{0.87}$ & 78.38$_{1.0}$ & 59.28$_{3.18}$ &  50.13$_{5.16}$ & 65.23\\ 
      \textbf{AdapterDrop$^{AA}$} &  42.66$_{7.02}$ &  69.52$_{1.03}$ & 74.15$_{2.19}$ & \bfseries 89.01$_{0.49}$ &  38.44$_{4.51}$ & 82.05$_{1.05}$ & 78.19$_{1.04}$ & 59.28$_{2.6}$ & 49.3$_{6.36}$ & 64.73\\
      \textbf{AdapterDrop$^{13}$} & 43.05$_{6.41}$ & 69.12$_{0.88}$ & 72.82$_{1.83}$ & 88.97$_{0.6}$ & 36.89$_{5.03}$ & 80.77$_{1.32}$ & 77.86$_{0.8}$ & 58.56$_{2.44}$ & 49.01$_{6.57}$ & 64.12 \\
      \textbf{AA-focused$^{spec}$} &  54.96$_{2.66}$ & \ 69.52$_{1.14}$ & \bfseries 77.30$_{1.27}$ & 87.94$_{1.10}$ & 39.51$_{3.47}$ & 84.30$_{0.69}$ & \bfseries 78.92$_{1.70}$ & 59.20$_{2.58}$ & 48.73$_{6.27}$ & 66.71 \\

      \textbf{AA-focused$^{uni}$} & \bfseries 56.13$_{1.88}$ & 69.32$_{2.29}$ & 76.85$_{2.37}$ & 87.89$_{1.47}$ & \bfseries 41.75$_{3.83}$ & 83.48$_{1.25}$ & 78.00$_{0.35}$ & \bfseries 60.42$_{1.75}$ & \bfseries 50.42$_{5.07}$ & \bfseries 67.14\\
      \textbf{AA-focused$^{sim}$} & 55.85$_{2.62}$ & \bfseries 69.86$_{2.56}$ & \bfseries 77.30$_{1.93}$ & 87.57$_{1.69}$ & 39.79$_{1.42}$ & 83.23$_{1.61}$ & 78.75$_{1.26}$ & 60.07$_{1.62}$ & 49.58$_{6.75}$ & 66.89\\
      \hline
    \textbf{|AA|} & 12.8$_{6.0}$ & 16.8$_{1.3}$ & 16.4$_{2.6}$ & 14.6$_{2.1}$ & 10.6$_{8.3}$ & 19.6$_{1.4}$ & 16.6$_{2.4}$ & 14.3$_{6.8}$ & 12.6$_{3.2}$ \\ 
      \textbf{|AA-focused$^{spec}$|} & 14 & 17 & 18 & 15 & 17 & 18 & 14 & 16 & 14\\

      \hline
    \multicolumn{11}{c}{\textbf{Full Data}} \\ \hline
    \textbf{Baseline} & 85.08 & 88.68 & \bfseries 91.95 &  93.00 &  58.28 & 89.75 & 83.12 & \bfseries 70.39 & 56.34 & \bfseries 79.62 \\
        \textbf{AA} & 84.73 & 88.38 & 91.01 & 92.55 & 57.60 & \bfseries 90.11 & 82.36 & 63.18 & 53.52 & 78.16 \\

    \textbf{AdapterDrop$^{AA}$} & 84.96 & \bfseries 88.75 & 91.38 & \bfseries 93.35 & \bfseries 58.63 & 89.85 & 82.84 & 66.06 & 56.34 & 79.12\\
    \textbf{AdapterDrop$^{13}$} & 84.73 & 87.15 & 90.92 & 92.78 & 57.42 & 88.84 & 83.34 & 64.25 & 56.34 & 78.42 \\
    \textbf{AA-focused$^{spec}$} & 84.77 & 88.46 & 91.38 & 92.32 & 56.79 & 89.74 & 83.42 & 64.98 & \bfseries 57.75 & 78.84 \\
    \textbf{AA-focused$^{uni}$}  & 85.41 &  88.61 & 91.51 & 92.66 & 54.62 & 89.34 & \bfseries 84.88 & 67.15 & 56.34 & 78.94 \\
    \textbf{AA-focused$^{sim}$}  & \bfseries 85.32 & 88.41 & 91.85 & 91.4 & 57.96 & 89.38 & 84.42 & 67.86 & \bfseries 57.75 & 79.37 \\
    \hline
    \textbf{|AA|} & 14 & 18 & 17 & 18 & 20 & 20 & 18 & 16 & 15 \\
    \textbf{|AA-focused$^{spec}$|} & 14 & 18 & 17 & 18 & 20 & 20 & 18 & 16 & 15 \\
    \end{tabular}
    }
    \caption{Comparing the results of (a) the standard adapter model that includes an adapter layer on all the 24 BERT-large layers (\emph{Baseline}), (b) adaptable adapter (\emph{AA}), (c) \emph{AdapterDrop}, and (d) \emph{AA-focused} adapters, in which the architecture of the adapter is selected based on the selected layers by \emph{AA}. The architecture of \emph{AA-focused$^{spec}$} is selected based on the selected layers by \emph{AA} for the corresponding task and data setting when the random seed is 42. The architecture of \emph{AA-focused$^{uni}$} is selected based on the selected layers by \emph{AA} for the task of \emph{QQP} on the \emph{Low-data-100} setting and for random seed 42. \emph{AA-focused$^{sim}$} only contains an adapter layer with a rational activation function at the last 13 layers of BERT-large, i.e., the total number of adapter layers in \emph{AA-focused$^{uni}$}. The number of layers at the inference time for the \emph{AdapterDrop$^{AA}$} experiments are selected based on the number of layers in the corresponding \emph{AA-focused$^{spec}$} experiments. The number of inference time layers for \emph{AdapterDrop$^{13}$} equals 13. Except for \emph{Full Data}, the reported results are averaged over five random seeds. The subscript reports the corresponding standard deviation. The \emph{Full Data} results are reported for one random seed. The \emph{|AA|} rows report the average number of selected adapter layers by \emph{AA} using different random seeds. \emph{|AA-focused$^{*}$|} rows report the number of added adapter layers in the corresponding \emph{|AA-focused$^{*}$|} experiments. \emph{|AA-focused$^{uni}$|} and \emph{|AA-focused$^{sim}$|} are the same for all data settings.  \emph{|AdapterDrop$^{*}$|} rows report the number of included adapter layers for the corresponding \emph{AdapterDrop} experiment at the inference time. \emph{|AdapterDrop$^{AA}$|} is always the same as the corresponding \emph{|AA-focused$^{spec}$|}, and \emph{|AdapterDrop$^{13}$|} is always the same as \emph{AA-focused$^{sim}$}.  The test data is the same for all the experiments. The \emph{Avg} column reports the average score across all datasets. The highest performances for each dataset and each data setting are boldfaced. } 
    \label{tab:glue_results_test_bert-large_main}
\end{table*}
}

\newcommand{\insertoverallResults}{
\begin{table*}[ht]
    \centering
     \resizebox{\textwidth}{!}{%
    \begin{tabular}{lllllllllll}
     & \textbf{MNLI} &	\textbf{QQP} &	\textbf{QNLI} &	\textbf{SST-2} &	\textbf{CoLA} &  \textbf{STS-B} & \textbf{MRPC} & \textbf{RTE} & \textbf{WNLI}  & \textbf{Avg}\\ \hline
     \textbf{Baseline} & \bfseries 83.53$_{0.19}$ & \bfseries 88.12$_{0.14}$ & \bfseries 90.63$_{0.26}$ & \bfseries 91.74$_{0.36}$ & \bfseries 56.51$_{0.84}$ & \bfseries 88.48$_{0.14}$ & 84.8$_{1.07}$ & 63.83$_{1.4}$ & \bfseries 54.08$_{6.64}$ & \bfseries 77.97\\
     \textbf{AA} & 82.89$_{0.43}$ & 88.09$_{0.16}$ & 89.96$_{0.25}$ & 91.31$_{0.51}$ & 51.44$_{1.82}$ & 88.25$_{0.17}$ & \bfseries 85.09$_{1.06}$ & \bfseries 64.25$_{1.72}$ & 52.11$_{7.61}$ & 77.05\\ \hline
    \textbf{AA-Layers} & 9.8$_{0.3}$ & 11.2$_{0.7}$ & 10.6$_{1.0}$ & 9.8$_{1.1}$ & 8.6$_{2.1}$ & 11.4$_{0.4}$ & 9.0$_{0.6}$ & 9.4$_{0.7}$ & 8.0$_{1.4}$ \\  \hline
    \multicolumn{10}{c}{\textbf{Low-data-100}} \\ \hline
      \textbf{Baseline} & 35.66$_{3.38}$ & 29.70$_{0.86}$ & 60.51$_{4.5}$ & 51.54$_{2.14}$ & -1.27$_{3.56}$ & 41.52$_{5.93}$ & \bfseries 74.86$_{0.12}$ & \bfseries 50.4$_{2.98}$ & 54.93$_{5.84}$ & 44.21\\
      \textbf{AA} & \bfseries 37.05$_{2.35}$ & \bfseries 30.59$_{0.68}$ & \bfseries 62.52$_{4.27}$ & \bfseries 52.73$_{2.55}$ & \bfseries -0.08$_{0.16}$ & \bfseries 48.73$_{23.91}$ & 74.83$_{0.07}$ & 50.18$_{3.21}$ & \bfseries 55.21$_{6.13}$ & \bfseries 45.75\\
     \hline
      \textbf{|AA|} & 6.4$_{1.8}$ & 8.6$_{2.1}$ & 8.8$_{1.7}$ & 8.6$_{1.6}$ & 7.4$_{2.4}$ & 10.8$_{0.7}$ & 9.4$_{1.4}$ & 9.4$_{1.4}$ & 8.2$_{0.9}$\\
\hline
    \multicolumn{10}{c}{\textbf{Low-data-300}} \\ \hline
     \textbf{Baseline} & 37.88$_{4.09}$ & 49.24$_{10.32}$ & 68.17$_{2.9}$ & 75.53$_{3.49}$ & 3.40$_{8.59}$ & 69.39$_{15.05}$ & 75.99$_{1.2}$  & 54.22$_{2.96}$ & \bfseries 47.61$_{4.91}$ & 53.49\\
     \textbf{AA} & \bfseries 40.27$_{4.78}$ & \bfseries 66.31$_{1.86}$ & \bfseries 74.03$_{2.03}$ & \bfseries 76.42$_{6.07}$ & \bfseries 3.56$_{5.49}$ & \bfseries 82.06$_{2.24}$ & \bfseries 76.12$_{0.89}$ & \bfseries 54.73$_{3.09}$ & 47.04$_{5.46}$ & \bfseries 57.84\\ \hline
     \textbf{|AA|} & 10.4$_{1.6}$ & 10.8$_{0.7}$ & 11.0$_{0.8}$ & 9.4$_{1.3}$ & 7.6$_{2.0}$ & 10.8$_{0.7}$ & 9.6$_{1.0}$ & 9.8$_{1.4}$ & 8.2$_{1.1}$ \\ \hline
    \multicolumn{10}{c}{\textbf{Low-data-500}} \\ \hline
     \textbf{Baseline} & 42.82$_{2.4}$ & 67.63$_{1.44}$ & 72.7$_{1.31}$ & 83.46$_{0.64}$ & \bfseries 20.9$_{4.14}$ & 81.97$_{0.89}$ & 76.51$_{0.95}$ & \bfseries 57.11$_{2.93}$ & \bfseries 52.11$_{6.96}$  & 61.69\\
      \textbf{AA} & \bfseries 47.72$_{1.67}$ & \bfseries 69.27$_{0.89}$ & \bfseries 75.649$_{1.9}$ & \bfseries 84.52$_{1.18}$ & 19.13$_{14.46}$ & \bfseries 83.74$_{0.67}$ & \bfseries 78.03$_{2.33}$ & 55.96$_{3.08}$ & 51.83$_{6.13}$ & \bfseries 62.87 \\ \hline
      \textbf{|AA|} & 9.8$_{1.1}$ & 10.4$_{1.3}$ & 10.0$_{0.8}$ & 9.2$_{0.7}$ & 9.4$_{1.8}$ & 10.6$_{1.4}$ & 9.8$_{1.6}$ & 9.6$_{1.0}$ & 8.0$_{1.5}$\\ \hline

    \end{tabular}
    }
    \caption{Comparing the results of (a) the baseline adapter model that includes an adapter layer on all BERT-base layers (\emph{Baseline}), and (b) the adaptable adapter (\emph{AA}). The reported results are averaged over five different random seeds. The subscript reports the corresponding standard deviation. \emph{|AA|} reports the average number of selected adapter layers by the adaptable adapter over different runs. The \emph{full data} results show the performance when the model is trained on all the available training data. The \emph{Low-data-$X$} settings report the results when only $X$ examples are used for training the model. The test data is the same for all the experiments.  }
    \label{tab:glue_results_test_bert}
\end{table*}

}

\newcommand{\insertLayerNumberSelection}{
\begin{table*}[!htb]
    \centering
     \resizebox{\textwidth}{!}{%
    \begin{tabular}{llllllllllll}
     \textbf{Adap. layers} & \textbf{MNLI} &	\textbf{QQP} &	\textbf{QNLI} &	\textbf{SST-2} &	\textbf{CoLA} &  \textbf{STS-B} & \textbf{MRPC} & \textbf{RTE} & \textbf{WNLI} & \textbf{Avg} \\ \hline
    \multicolumn{11}{c}{\textbf{Low-data-300}} \\ \hline
     \textbf{13} & 45.97$_{2.08}$ & 68.36$_{1.36}$ & 73.98$_{2.68}$ & 86.83$_{1.9}$ & 37.43$_{3.1}$ & 78.81$_{3.58}$ & 76.66$_{1.3}$ & 55.96$_{2.81}$ & 48.44$_{5.53}$ & 63.61\\
     \textbf{12} & 36.84$_{5.51}$ & 62.43$_{5.65}$ & 65.77$_{3.43}$ & 84.63$_{3.64}$ & 13.23$_{12.63}$ & 77.08$_{2.36}$ & 75.27$_{0.39}$ & 54.30$_{3.75}$ & 46.76$_{5.45}$ & 57.37\\
     \textbf{11} & 36.16$_{5.12}$ & 62.59$_{5.8}$ & 67.93$_{1.42}$ & 79.95$_{14.16}$ & 16.32$_{11.65}$ & 73.22$_{6.75}$ & 76.42$_{1.19}$ & 56.53$_{2.02}$ & 46.2$_{4.12}$ & 57.26\\

    \end{tabular}
    }
    \caption{Evaluating the impact of the number of adapter layers on the overall performance. The adapter layers are added to the top $n$ layers of the model for $n=13, 12, 11$. Adapter layers contain rational activation, i.e., $n=13$ is equivalent to \emph{AA-focused$^{sim}$}. Results are reported for the \emph{low-data-300} setting.}
    \label{tab:number_layers}
\end{table*}
}

\newcommand{\insertAblation}{
\begin{table*}[!htb]
    \centering
     \resizebox{\textwidth}{!}{%
    \begin{tabular}{llllllllllll}
     & \textbf{MNLI} &	\textbf{QQP} &	\textbf{QNLI} &	\textbf{SST-2} &	\textbf{CoLA} &  \textbf{STS-B} & \textbf{MRPC} & \textbf{RTE} & \textbf{WNLI} & \textbf{Avg} \\ \hline
    \multicolumn{11}{c}{\textbf{Low-data-300}} \\ \hline
     \textbf{Baseline} & 36.55$_{4.76}$ & 61.50$_{8.66}$ & 69.62$_{1.24}$ & 79.86$_{14.15}$ & 30.40$_{5.48}$ & 78.24$_{2.81}$ & 76.55$_{1.31}$ & 51.62$_{3.21}$ & 45.92$_{4.33}$ & 58.91 \\
    \textbf{AA} & 37.14$_{2.49}$ & 66.07$_{0.38}$ & 71.33$_{1.82}$ & 72.52$_{16.59}$ & 26.05$_{8.74}$ & \bfseries 82.08$_{1.6}$ & 74.03$_{2.21}$ & 51.83$_{2.84}$ &  47.04$_{4.76}$ & 58.68 \\
    \textbf{Switch-Only} &  35.05$_{2.81}$ & 43.8$_{16.02}$ & 65.59$_{2.61}$ & 61.86$_{6.26}$ & 9.77$_{12.86}$ & 75.41$_{3.29}$ & 75.37$_{0.7}$ & 50.18$_{3.44}$ & 45.92$_{3.03}$ & 51.44\\
     \textbf{Rational-Only} & 37.72$_{3.88}$ & 64.75$_{2.51}$ & 69.69$_{1.04}$ & 79.86$_{14.15}$ & 23.20$_{8.33}$ & 78.58$_{1.94}$ & 75.84$_{1.07}$ & 52.27$_{3.11}$ & 46.48$_{3.88}$ & 58.70\\
    \textbf{Baseline$^{13}$} & 37.98$_{5.80}$ & 63.37$_{4.72}$ & 68.76$_{1.55}$ & 85.16$_{3.63}$ & 12.11$_{12.69}$ & 77.96$_{2.23}$ & 75.25$_{0.71}$ & 54.44$_{2.06}$ & 45.35$_{3.72}$ & 57.80 \\
      \textbf{AA-focused$^{sim}$} & 45.97$_{2.08}$ & 68.36$_{1.36}$ & 73.98$_{2.68}$ & 86.83$_{1.90}$ & \bfseries 37.43$_{3.10}$ & 78.81$_{3.58}$ & \bfseries 76.66$_{1.30}$ & 55.96$_{2.81}$ & 48.44$_{5.53}$ & 63.61 \\

     \hline
     \textbf{|AA|} & 17.0$_{1.3}$ & 16.2$_{1.0}$ & 14.8$_{1.8}$ & 12.8$_{3.2}$ & 16.8$_{2.2}$ & 18.6$_{1.9}$ & 16.0$_{1.1}$ & 12.4$_{1.2}$ & 12.4$_{2.1}$  \\
     \textbf{|Switch-Only|} & 14.0$_{1.1}$ & 15.8$_{2.5}$ & 17.0$_{1.9}$ & 16.2$_{2.8}$ & 16.4$_{1.9}$ & 16.4$_{1.5}$ & 17.8$_{1.7}$ & 15.0$_{2.1}$ & 14.0$_{1.7}$ &  \\

    \end{tabular}
    }
    \caption{Evaluating the impact of rational in adaptable adapters. Experiments are run for five different random seeds. \emph{Switch-only} shows the results  when learnable switches are used with standard adapter layers, i.e., linear layers with the ReLU activation. \emph{Rational-only} shows the result when all the activation functions in the standard adapter are replaced with rational. \emph{Baseline$^{13}$} contains a standard adapter layer on the last 13 transformer layer. \emph{AA-focused$^{sim}$} contains adapter layers with rational activation on the last 13 layers. }
    \label{tab:ablations}
\end{table*}
}

\clearpage{}%

\title{Adaptable Adapters}

\author{%
Nafise Sadat Moosavi$^1$\thanks{The work has been mostly carried out during the employment at the UKP Lab, TU Darmstadt.}, Quentin Delfosse$^3$, Kristian Kersting$^{2,3}$, Iryna Gurevych$^{2,4}$\\ \\
$^1$ Department of Computer Science, The University of Sheffield \\
$^2$ Hessian Center for AI (hessian.AI) \\
$^3$ AI \& Machine Learning Lab, %
$^4$ Ubiquitous Knowledge Processing Lab (UKP Lab), \\
        Department of Computer Science, Technical University of Darmstadt \\ 
        \url{https://www.ukp.tu-darmstadt.de}
}

\date{}

\begin{document}
\maketitle
\begin{abstract}
State-of-the-art pretrained NLP models contain a hundred million to trillion parameters.
Adapters provide a parameter-efficient alternative for the full finetuning in which we can only finetune lightweight neural network layers on top of pretrained weights.
Adapter layers are initialized randomly. However, existing work uses the same adapter architecture---i.e., the same adapter layer on top of each layer of the pretrained model---for every dataset, regardless of the properties of the dataset or the amount of available training data.
In this work, we introduce adaptable adapters that contain (1) learning different activation functions for different layers and different input data, and (2) a learnable switch to select and only use the beneficial adapter layers.  We show that adaptable adapters achieve on-par performances with the standard adapter architecture while using a considerably smaller number of adapter layers. In addition, we show that the selected adapter architecture by adaptable adapters transfers well across different data settings and similar tasks. We propose to use adaptable adapters for designing efficient and effective adapter architectures. The resulting adapters (a) contain about 50\% of the learning parameters of the standard adapter and are therefore more efficient at training and inference, and require less storage space, and (b) achieve considerably higher performances in low-data settings.\footnote{The code will be available at \url{https://github.com/UKPLab/adaptable-adapters}.}   
\end{abstract}

\section{Introduction}
\label{sect:intro}

Recent improvements in NLP are heavily skewed towards using larger pretrained models \citep{roberts-etal-2020-much} and given their considerably better performances, using them is becoming unavoidable \citep{kaplan2020scaling}. 
Their improvements, however, come at the cost of significant computational resources at training and inference times. 
For instance, the number of parameters in recent pretrained models can vary from 110M in BERT-base \citep{devlin-etal-2019-bert} to 11 billion in T0 \citep{sanh2021multitask} to trillion parameters in Switch Transformers \citep{fedus2021switch}.
Using such models for each downstream application requires a vast amount of storage, training, and inference computation budget that is not accessible to every user.  

Instead of fine-tuning these massive numbers of parameters for each downstream task, we can use adapter architectures \citep{houlsby2019parameter,pfeiffer2020AdapterHub}.
Adapters are lightweight neural network layers that are added on top of each layer of the pretrained model. As opposed to the standard model fine-tuning, in which all layers are fine-tuned for the target task, adapter-based tuning freezes the transformer layers and only trains the newly added adapter layers.
Since the majority of parameters---i.e., the layers of the large pretrained model---are shared between different downstream tasks, the use of adapters results in parameter-efficient transfer learning.
In addition to their parameter-efficiency, \citet{he-etal-2021-effectiveness} show that training adapter-layers (a) outperforms fine-tuning the whole model  on low-data and cross-lingual settings, and (b) is more robust to overfitting. 

Existing work suggests that (a) different layers of the pretrained models may capture different aspects of the form, syntax, or meaning of the input text \citep{tenney-etal-2019-bert,clark-etal-2019-bert}, and (b) they may not be all needed for performing a given task \citep{houlsby2019parameter,Fan2020Reducing,ruckle-etal-2021-adapterdrop}.
In addition, adapter layers are initialized randomly. Therefore, it is not necessary to use the same adapter architecture for different downstream tasks and given different amounts of annotated data. 
However, existing works use the same adapter architecture for all the different input data, i.e., (a) one adapter layer on top of all the pretrained layers while using all the layers may not be necessary, and (b) the same activation function for all the layers and different tasks while the best activation function may vary for different tasks \citep{delfosse2021recrat}.

In this paper, we propose a systematic approach for designing more adequate and flexible adapter architectures by introducing the adaptable adapter (\emph{AA}).
Adaptable adapters (1) use a learnable activation function---called Rational activation \citep{molina2019pad}---instead of a constant activation in adapter layers allowing the adapter model to learn different activation functions at different adapter layers and for different tasks, and (2) consist of a learnable switch at each adapter layer to determine the beneficial adapter layers during training and to only use the selected layers during inference.

We evaluate adaptable adapters on the GLUE benchmark \citep{wang-etal-2018-glue} that consists of various text classification tasks. We perform evaluations based on different data settings in which different amounts of annotated examples are available for training.
Our results show that adaptable adapters achieve on-par performances with the full adapter architecture while using considerably fewer adapter layers at the inference.

We further propose to use adaptable adapters for designing efficient adapter architectures---i.e., to only add an adapter layer to the layers that are selected by the adaptable adapter.
We show that while the selected adapter architecture by \emph{AA}, called \emph{AA-focused}, is considerably more efficient at both training and inference times and requires less storage, it achieves on-par performances with the full adapter architecture when trained on all available training data and considerably outperforms it on low-data settings.
In addition, we show that the selected adapter architecture by \emph{AA} transfers well across similar tasks and different data settings. Therefore, we can train \emph{AA} using a limited amount of training data, and for one of the tasks, and then use the resulting \emph{AA-focused} architecture for different data settings and other similar tasks.

Overall, the contributions of this paper are as follows:
\begin{itemize}
    \item We propose adaptable adapters that introduce flexibility in adapter architectures by (a) selecting the beneficial adapter layers to use, and (b) learning the suitable activation function for each layer and each task.
    \item We propose to use adaptable adapters to design efficient adapters that require less training time, inference time, and storage space.
    \item We show that using fewer adapter layers with a learnable activation function considerably improves the performance on low-data settings.
\end{itemize}

\section{Related Work}

\subsection{Rational Activation}
\label{sect:rational}
Rational activation functions, empirically introduced as Pad\'{e} Activation Units \citep{molina2019pad}, are learnable activation functions that can approximate common activation functions as well as learn new ones.
The rational activation function $R(x)$ of order $m,n$ is defined as follows:
\begin{equation}
    R(x) =  \frac{\sum_{j=0}^m a_j x^j}{1+\left|\sum_{k=1}^n b_k x_k\right|} \nonumber
\end{equation}
where $a_j$ and $b_k$ are learnable parameters. 
These rational functions use an absolute value in the denominator to avoid potential poles, which will make the training unstable. Such rational activation functions provide stable training, as empirically shown in image classification and reinforcement learning \citep{molina2019pad, delfosse2021recrat}.
$R(x)$ can be initialized to initially approximate any of the known activation functions or with constant functions. \citet{molina2019pad} show that rationals outperform other commonly used activation functions in common image classification tasks. 
Rational activation functions are also integrated in Generative Adversarial Networks \citep{BoulleNT20}.
\citet{delfosse2021recrat} show that some of the layers in very deep pretrained Residual Networks tend to approximate activation functions' behavior, and we can achieve on-par or better performances with the full network by replacing some of the complete layers with rational activation functions.
Similar to this observation, as we show in \S~\ref{sect:rational_impact}, using rational activation functions instead of a constant activation (ReLU) in adapters allows them to achieve high accuracy using a fewer number of adapter layers.

\subsection{Reducing Model's Size for Efficiency}
Improving the efficiency of large pretrained models has received particular attention for the inference time.
The argument is that the effect of training cost is limited, i.e., the model can be trained once but it will be used many times. However, the inference time has a wide impact on the everyday use of NLP models. 

Existing approaches for improving the inference-time efficiency belong to two different categories: (a) the distillation and pruning techniques that create a smaller model for inference but often require re-training or fine-tuning the smaller model \citep{tang2019distilling,distilbert,voita-etal-2019-analyzing,sun-etal-2020-mobilebert,bai-etal-2021-binarybert}, and (b) on-demand network size reduction at the inference time.\footnote{There is another category that requires changes in the models' architectures. However, it would require re-training the large model. E.g., \citet{sukhbaatar-etal-2019-adaptive} propose new attention mechanisms that can process larger context with  no additional computational or memory costs.}
There are two different approaches in the second category, namely layer dropping and early exiting.

\citet{Fan2020Reducing} use layer dropping during the training that randomly drops the model's layers to make the model robust to the inference time layer selection. They show that it is possible to select sub-networks of any depth from large models at inference with limited impact on the performance and without the need for additional finetuning.
Layer dropping was previously investigated by \citet{huang2016deep} who propose to drop layers during training for regularizing the model and reducing the training time of deep convolutional networks.
\citet{ruckle-etal-2021-adapterdrop} use layer dropping for adapter architectures. They show that by randomly dropping adapter layers during training, they can prune the adapter model on-demand at the inference time.

\citet{schwartz-etal-2020-right} propose to add an output layer to each transformer layer. At inference time, while the model calculates the layer-wise representation, from the bottom layer to the top layer, it also makes the prediction using the associated classification layer. They use the output labels' scores of the classification layers as confidence scores to decide whether to exit early if the classifier is confident or to proceed to process the input with the next layers.
This hierarchical architecture offers an inference time-accuracy tradeoff by setting the confidence threshold.
The early exiting approach is similar to layer dropping in which the dropped layers are always from the last top layers. 

All these approaches select the number of layers to drop and the dropped layers heuristically at the inference time with the goal of improving the inference time.
Instead, the adaptable adapter is a systematic approach for selecting the useful adapter layers for the given task during training. Besides layer selection, an adaptable adapter allows for learning the desired activation function for different inputs.
As we show, we can use adaptable adapters to design efficient adapter architectures with a considerably smaller number of training parameters with on-par or considerably higher performances, especially with larger models and in low-data settings.  

\section{Proposed Architecture}

\subsection{Learnable Activation}
Empirical observations of performances have led experts in several fields to use different activation functions for different tasks. Functions from the ReLU family are usually used for neural network-based visual computing, Tanh has been used in PPO for reinforcement learning, while GeLU has progressively been adopted in transformers. 
With the growth of the models, and the complexity of the tasks they are applied to, choosing one fixed activation function to equip the complete architecture is suboptimal. 
By using rational (\S~\ref{sect:rational}), we let the adapter layer learn the suitable activation function at each different adapter layer, task, and dataset. 
In adaptable adapters, we replace the constant activation function of each adapter layer---i.e., ReLU in the default configuration used in AdapterHub \citep{pfeiffer2020AdapterHub}---with rational.

\begin{figure}[!htb]
     \centering
     \begin{subfigure}[b]{0.45\columnwidth}
         \centering
         \includegraphics[width=\textwidth]{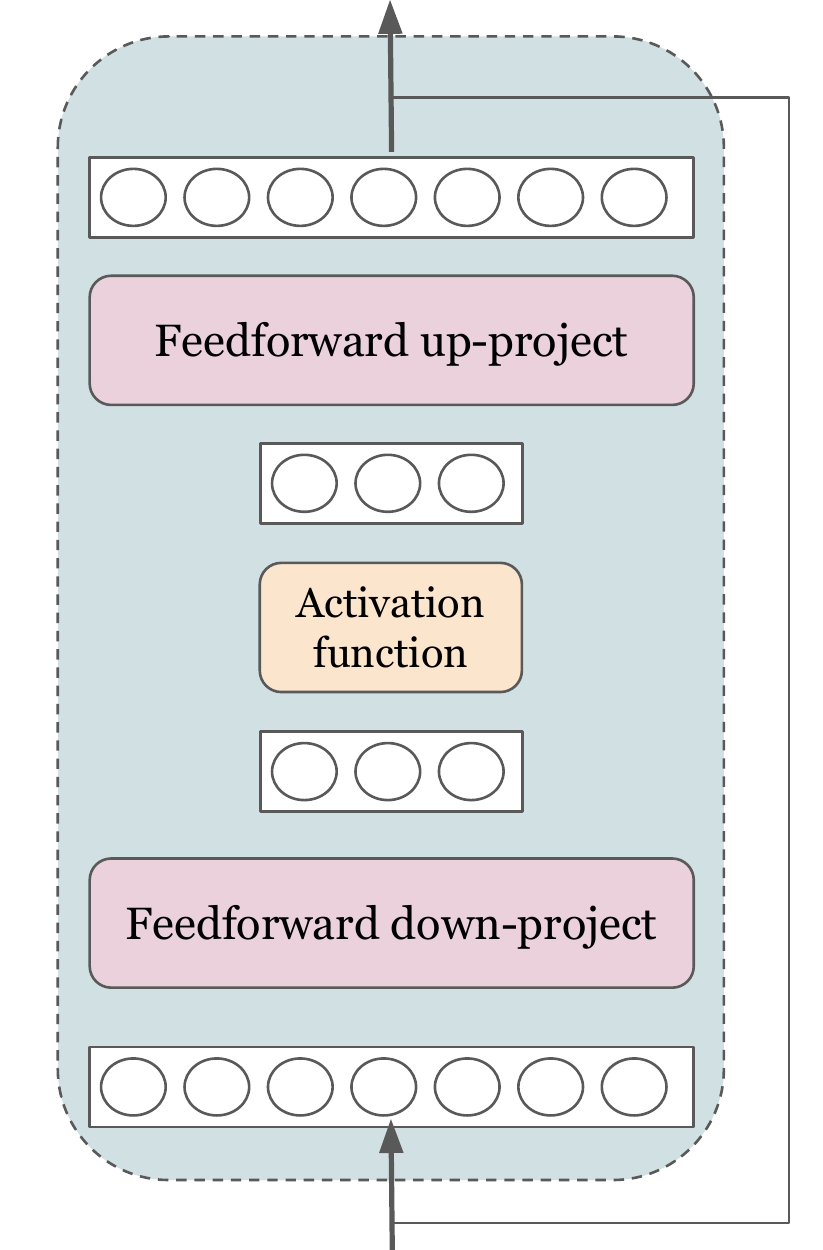}
         \label{fig:pfeiffer}
         \caption{}
     \end{subfigure}
     \hfill
     \begin{subfigure}[b]{0.45\columnwidth}
         \centering
         \includegraphics[width=\textwidth]{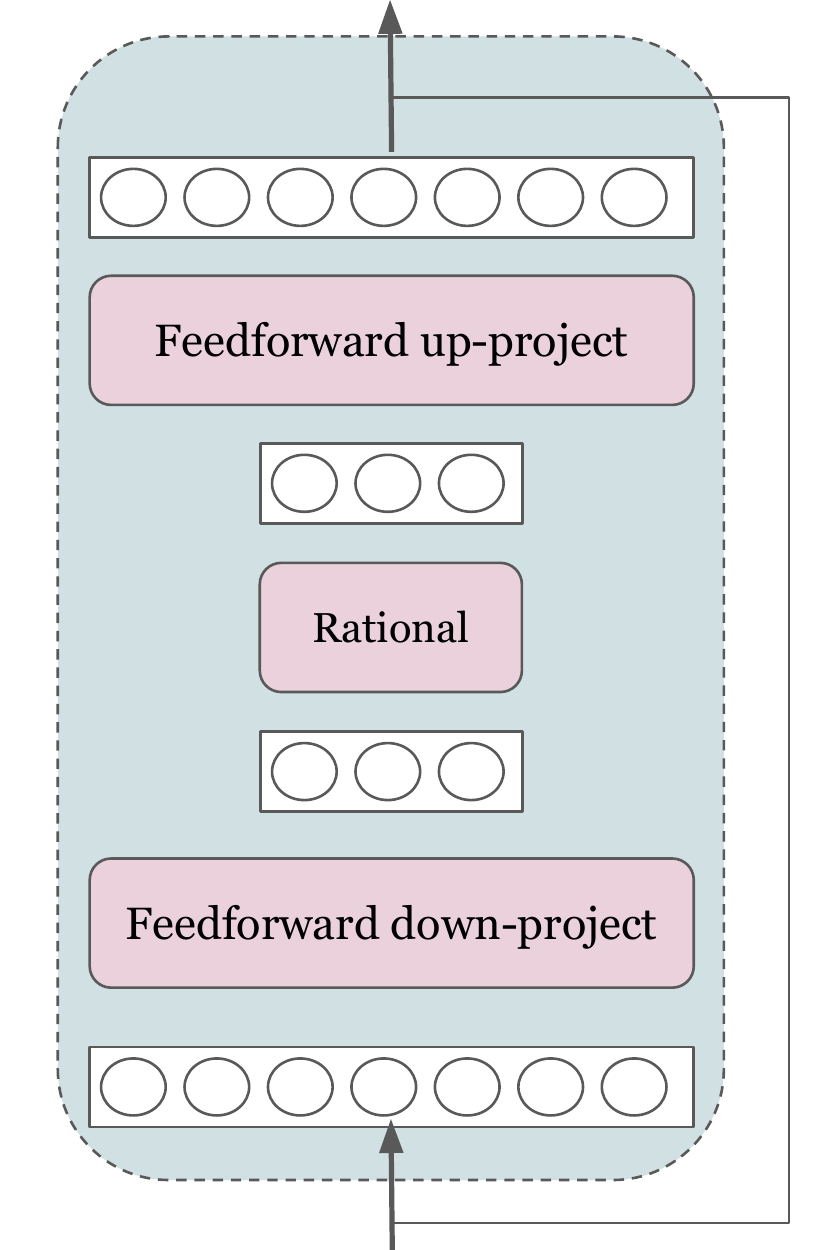}
         \label{fig:linear_rational}
        \caption{}
     \end{subfigure}
        \caption{(a) a standard adapter layer with linear feedforward layers and a fixed activation, (b) an adapter layer in adaptable adapters with linear feedforward layers and a rational activation. Learnable parameters are shown within pink boxes.}
        \label{fig:adapter_layers}
\end{figure}

Figure~\ref{fig:adapter_layers} shows a standard adapter layer as well as an adapter layer in adaptable adapters.

\subsection{Learnable Layer Selection}
\label{sect:switch}
\citet{houlsby2019parameter} examined various choices of adapter architectures. They report that using two feedforward linear layers---one down-project and one up-project layer---results in good performances while only introducing a few parameters.
Assuming $d$ is the dimensionality of the input---i.e., the embedding size of the transformer layer---the down-project layer maps the input dimension to $n$ where $n<d$, and the up-project layer maps the input dimension back to $d$. $n$ is called the hidden size of the adapter.
Each adapter contains a skip-connection that lets an adapter layer approximate an identity function, i.e., to pass the input of a transformer layer unchanged to the next layer.
The learnable switches in adaptable adapter explicitly model the selection between the feedforward adapter layer and the identity function. By examining the switch probabilities we can determine the adapter layers that are beneficial for the overall performance of the model.

As mentioned in \S~\ref{sect:intro}, existing work shows that different layers of the pretrained models capture different aspects of the input data, and not all of them are necessary for performing various tasks.
Therefore, for different input data, different layers may be of different importance. 
Adding a learnable switch at each adapter layer provides a more systematic approach to determining the beneficial layers for each input task during training.
We use the Gumbel Softmax ($\mathcal{GS}$) estimator as an end-to-end differentiable switch (hard attention) to make the network attend to an element of a set. 
Assuming $\pi_i$ are the probabilities of selecting each element of the set, i.e., $\forall_i \pi_i \geq 0, \sum_i \pi_i=1$, $\mathcal{GS}$ estimates the hard attention $y_i$ as follows:
\begin{equation}
    y_i = \frac{exp((log(\pi_i)+g_i)/\tau)}{\sum_j exp((log(\pi_j)+g_j)/\tau)} \nonumber
\end{equation}
where $g_i$ are i.i.d. samples from a Gumbel distribution, and $\tau$ is a temperature parameter. Setting $\tau$ to small values results in distributions that are similar to categorical ones.

\begin{figure}[!htb]
    \centering
    \includegraphics[width=0.5\columnwidth]{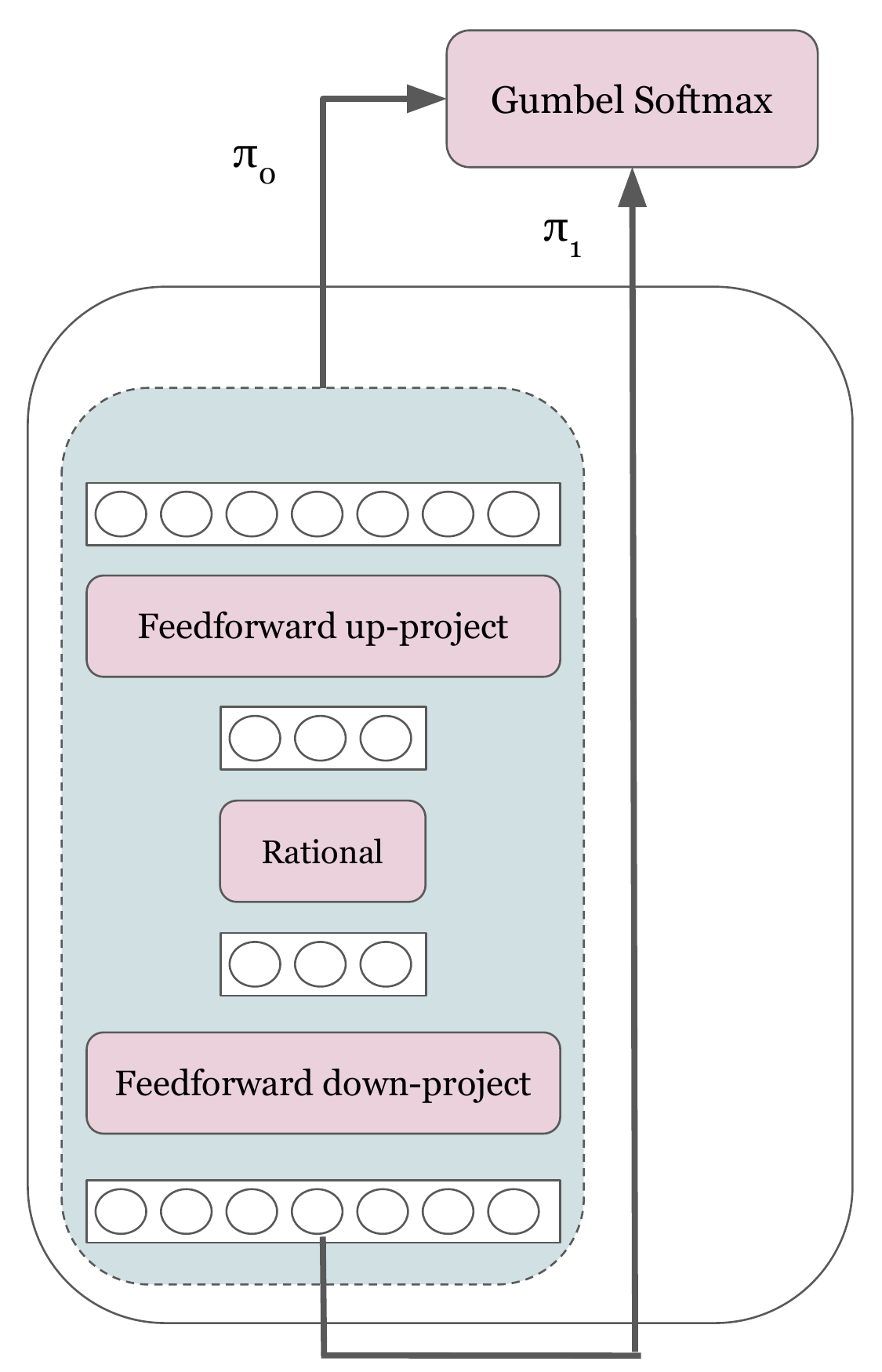}
    \caption{The adaptable adapter layer that consist of a Gumbel Softmax to choose between an adapter layer with a rational activation and an identity function.}
    \label{fig:gs}
\end{figure}

\subsection{Adaptable Adapters}

The adaptable adapter (\emph{AA}) is the combination of the learnable layer selection and the learnable activation function.
The learnable layer selection---i.e., a Gumbel Softmax estimator---selects between an adapter layer, with no skip connection, and an identity function with zero parameters that passes the input without any changes to the next layer.
The adapter layers in adaptable adapters consist of two linear layers---i.e., down-project and up-project layers---, and the non-linearity function between these two linear layers consists of a rational activation function.
The adaptable adapter allows to learn different adapter architectures for different input data by (a) learning to use a subset of adapter layers, and (b) learning a potentially different activation function at each layer.
Figure~\ref{fig:gs} shows the structure of an adapter layer in adaptable adapters.

\section{Experimental Setup}
\subsection{Datasets}
We use the English text classification datasets from the GLUE benchmark \citep{wang2018glue} including MNLI \citep{williams-etal-2018-broad}, QQP\footnote{\url{https://www.quora.com/profile/Ricky-Riche-2/First-Quora-Dataset-Release-Question-Pairs}}, QNLI \citep{rajpurkar-etal-2016-squad}, SST-2 \citep{socher-etal-2013-recursive}, CoLA \citep{warstadt-etal-2019-neural}, STS-B \citep{cer-etal-2017-semeval}, MRPC \citep{dolan-brockett-2005-automatically}, RTE \citep{dagan2006pascal}, and WNLI \citep{levesque2011winograd}.
Table~\ref{tab:glue_data} shows the number of training examples and the evaluation metric for each dataset.

\insertGlueSpec

\subsection{Transformer Model} 
As the base model, we use the BERT-large model \citep{devlin-etal-2019-bert}. BERT-large contains 24 layers, an embedding size of 1024, and a total number of 340M parameters.\footnote{The results for BERT-base are reported in the appendix. BERT-base contains 12 layers, an embedding size of 768, and 110M parameters.}

\subsection{Adapter Models}
\paragraph{Baseline} As a baseline adapter, we use the adapter layers with the pfeiffer configuration from AdapterHub \citep{pfeiffer2020AdapterHub}.
The adapter layers with the pfeiffer configuration are similar to the one in  Figure~\ref{fig:adapter_layers}, in which learnable parameters include two feedforward layers. For BERT-base, each pfeiffer layer consists of 73.7k parameters\footnote{The reduction factor in the down-project layer is 16 which results in (768/16) x 768 x 2 parameters for each adapter layer.} resulting in a total number of 884.7K.
For BERT-large, the number of parameters for each adapter layer is 131K, and the total number of parameters is 3.1M. We see that as the underlying model gets larger, the number of parameters in adapters also increases notably. Therefore, adapter architecture selection using \emph{AA} is a potential solution to control this exponential increase to some extent. 

\paragraph{Adaptable Adapter (AA)}
For the rational activation, similar to \citet{molina2019pad}, we use order $m=5$ and $n= 4$ for rational. Therefore, the rational activation function only consists of ten learnable parameters. The rational activation can be initialized to initially estimate an existing function. Based on our preliminary experiments, using $f(x)=1$ for initializing $R(x)$ results in better performances on the GLUE benchmark.

For the Gumble-Softmax switch, we set the temperature parameter $\tau$ to 0.1, and we initialize $\pi_i$ to 0.5 for both inputs---i.e., the same initial probability for the rational adapter and the identity function. 

\paragraph{AA-focused} We can use the selected architecture by \emph{AA} for designing a new adapter architecture, i.e., to only include an adapter layer---with a rational function---at layers in which the switch has selected the adapter layer over the identity function. We call this architecture \emph{AA-focused}.
Note that compared to \emph{AA}, \emph{AA-focused} is more efficient both at training and inference time, as it includes a fewer number of layers and no switch functions. It also requires less storage space for saving the new adapter weights. Also, training \emph{AA} includes both the architecture selection and training of the adapter layers, which are initialized randomly, simultaneously. As a result, as we see in our evaluations, \emph{AA-focused} achieves higher performances as its training is only focused on training the adapter layers.

\paragraph{AdapterDrop \citep{ruckle-etal-2021-adapterdrop}} During training, AdapterDrop randomly drops the first $n$ layers in which $n$ varies for different iterations. At inference, $n$ can be set to any desired number of layers.  
In our experiments, we select $n$ based on the number of dropped layers by \emph{AA}, i.e., the number of layers that are not selected by the switch functions.

\subsection{Experiments}
We evaluate the models in different settings: (a) using full training data, and (b) low-data settings.
For all the experiments, we consider 25\% of the training data as the development set and use the official development sets as the test data.  
We perform the low-data evaluations when 100, 300, and 500 annotated examples are available.\footnote{Selected training examples for low-data experiments are the same for all models given the same random seed.} The test data is the same for all the evaluations.
We run all the low-data experiments for 20 epochs and five different random seeds\footnote{42, 92, 111, 245, and 651.}. We report the average and standard deviation over the five different runs. When training on full datasets, the experiments are computationally very expensive using BERT-large. Therefore, for this setting, we only report the results using the first random seed. All experiments are done on one A100 NVIDIA GPU.
All implementations are based on AdapterHub \citep{pfeiffer2020AdapterHub}.%

\insertArchitectureSelectionCrossDataRegime

\section{Evaluation}
Table~\ref{tab:glue_results_test_bert-large_main} presents the results of \emph{Baseline}, \emph{AdapterDrop}, \emph{AA}, and \emph{AA-focused}.
\emph{AA} selects different layers for different tasks and different random seeds.\footnote{For instance, the selected layers for \emph{RTE} are as follows for different runs of \emph{Low-data-100}: \{0, 2, 5, 11, 12, 13, 16, 17\}, \{3, 4, 5, 6, 7, 8, 9, 10, 12, 13, 15, 19, 21\}, \{2, 3, 4, 6, 9, 12, 14, 16, 17, 18, 20, 22, 23\}, \{0, 2, 6, 8, 9, 11, 13, 14, 17, 19, 23\}, \{1, 2, 5, 10, 11, 14, 16, 20, 21, 22, 23\}.}
We evaluate three configurations for \emph{AA-focused}:
\begin{itemize}
    \item \textbf{AA-focused$^{{spec}}$}: for each task, we design the corresponding \emph{AA-focused} based on the selected architecture by \emph{AA} for that task given and the first random seed (42). For instance, the \emph{AA-focused} architecture is the same for all the experiments of \emph{RTE} for \emph{Low-data-100}---i.e., over the five different random seeds---. However, it is different for the rest of the tasks and different data settings.
    \item \textbf{AA-focused$^{{uni}}$}: we design this adapter architecture of all tasks and data settings based on a single random seed, single task, and a single data regime, i.e.--- random seed 42, the \emph{QQP} task, and \emph{low-data-100}. We choose \emph{low-data-100} because the architecture selection process---i.e., training \emph{AA}---is very fast in this setting. We select the selected architecture by \emph{QQP} because \emph{AA} selects the smallest number of layers for \emph{QQP} when the random seed is 42. The selected layers are \{2, 6, 10, 12, 14, 15, 16, 18, 19, 20, 21, 22, 23\}, i.e., three layers from the first half of the original 24 layers, and ten layers from the second half. The results of \emph{{AA-focused$^{{uni}}$}} compared to {AA-focused$^{{spec}}$} indicate whether the selected architecture by \emph{AA} transfers between similar tasks and different data settings.  
    \item \textbf{AA-focused$^{{sim}}$}: we design a simplified adapter based on \emph{AA} in which we only use the number of selected layers, instead of the layer numbers, in a single random seed, single task, and a single data setting. We use the number of selected layers when the random seed is 42 for the \emph{QQP} task and the \emph{low-data-100} setting, i.e., 13. As investigated by \citet{houlsby2019parameter}, the last adapter layers are in general more effective. As a result, we add adapter layers, with rational activation, to the last 13 transformer layers in \emph{AA-focused$^{{sim}}$} experiments. The results of \emph{{AA-focused$^{{sim}}$}} compared to {AA-focused$^{{uni}}$} show whether only the number of selected layers by \emph{AA} matters or it is also important to specify at which layers to add the adapters.
\end{itemize}

The number of inference layers for \emph{AdapterDrop$^{AA}$} are equivalent to the number of layers in \emph{AA-focused$^{{spec}}$} experiments for each task and data setting.
The number of layers for \emph{AdapterDrop$^{13}$} is 13, which is the same as \emph{AA-focused$^{{uni}}$} and \emph{AA-focused$^{{sim}}$}.
Note that the number of layers for \emph{AA-focused} experiments are the same both at training and inference while it is not the case for \emph{AdapterDrop}.

 The \emph{|AA|} rows in Table~\ref{tab:glue_results_test_bert-large_main} show the average number of selected layers for each task over the five different random seeds. 
\emph{|AA-focused$^{*}$|} rows report the number of added adapter layers in the corresponding \emph{AA-focused$^{*}$} experiments. \emph{|AdapterDrop$^{*}$|} rows report the number of included adapter layers for the corresponding \emph{AdapterDrop} experiments at the inference time. 

\insertLayerNumberSelection
\insertAblation

We make the following observations from the results of Table~\ref{tab:glue_results_test_bert-large_main}:
\begin{itemize}
    \item \emph{AA} achieves on-par performances with the \emph{Baseline}, and on average it uses about 13-15 layers out of 24 layers. We can use this insight for designing efficient adapter architectures.
    \item All \emph{AA-focused} architectures considerably outperform \emph{Baseline} in all the the tasks in low-data settings while using considerably smaller number of parameters, and therefore, being considerably more efficient. For instance, while \emph{AA-focused$^{uni}$} only uses 13 layers out of 24 layers---i.e., reducing the number of training parameters from 3M to 1.7M---, it outperforms the \emph{Avg} score by 4.24, 5.57, and 2.35 points in \emph{Low-data-100}, \emph{Low-data-300}, and \emph{Low-data-500}, respectively.     
    \item The high performances of \emph{AA-focused$^{uni}$} show that the selected architecture by \emph{AA} for one task and one data setting transfers well to other data regimes and similar tasks.\footnote{It even outperforms \emph{AA-focused$^{spec}$} showing that \emph{AA-focused$^{spec}$} may have overfitted to the development sets. We have not performed hyperparameter selection for our experiments. Using better hyperparameters may improve the results of different settings.} Therefore, it is not necessary to design the adapter architecture separately for a different amount of available data and similar tasks.

    \item \emph{AA-focused$^{sim}$} and \emph{AdapterDrop$^{13}$} both use the last 13 adapter layers during the inference while the results of \emph{AA-focused$^{sim}$} are considerably higher for all data regimes. This indicates the importance of rational activation in adaptable adapters. We will further investigate the impact on rational activation in the next section.
    \item In average, \emph{AdapterDrop$^{AA}$} contains more inference layers compared to \emph{AdapterDrop$^{13}$}. However, there is not a significant difference between their performances. They achieve on-par or lower results compared to \emph{Baseline}.
\end{itemize}

\paragraph{Evaluating the impact of AA on selecting the number of beneficial layers.}
In the results of Table~\ref{tab:glue_results_test_bert-large_main}, we select the number of layers in \emph{AA-focused$^{sim}$}, i.e., 13 , based on the minimum number of selected layers by \emph{AA} on the \emph{low-data-100} setting and for random seed 42.
\emph{AA-focused$^{sim}$} is equivalent to an adapter architecture in which only the last 13 adapter layers are added to the model.
To investigate whether the improvements of  \emph{AA-focused$^{sim}$} over the baseline are only due to using a fewer number of adapter layers, we report the results of an adapter architecture in which only the last $n$ adapter layers are added to the model, e.g., for $n=13$ the resulting architecture is the same as \emph{AA-focused$^{sim}$}.
Table~\ref{tab:number_layers} shows the result of this experiment for $n=13, 12, 11$. We observe that by decreasing the number of layers from 13 to 12, the overall performance drops notably from 63.61 to 57.37. 

\begin{figure*}[!htb]
    \centering
    \includegraphics[width=.95\textwidth]{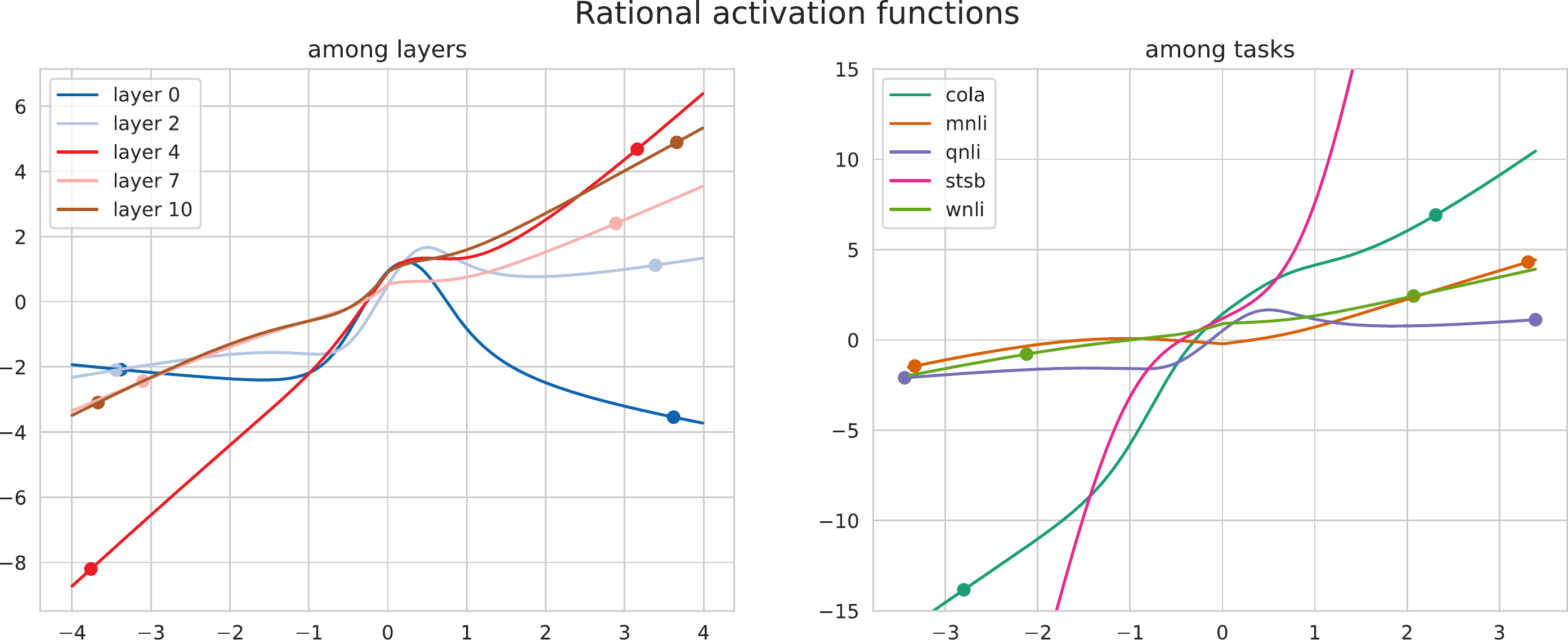}
    \caption{Learned rational activation functions differ according to their place within the network and to the task they are trained for. Right: activation functions at different layers within adapters trained on the QNLI task. Left: activation functions trained at layer 2 of adapters trained on different tasks.}
    \label{fig:gs}
\end{figure*}

\paragraph{Evaluating the impact of rational activation.}
\label{sect:rational_impact}
The results of \emph{AA-focused} experiments vs. \emph{Baseline} in Table~\ref{tab:glue_results_test_bert-large_main} mostly emphasize the impact of layer selection by the learnable switches in \emph{AA}. In this section, we investigate the impact of learnable activation functions in more details in the evaluations of Table~\ref{tab:ablations}.

First, we replace all rationals in \emph{AA} with ReLU. The results are reported in the \emph{Switch-Only} row. By comparing the results of \emph{AA} and \emph{Switch-only} we observe that the use of rational activation considerably improves the performance of \emph{AA}, i.e., using rational is a key component to achieving higher performances with fewer layers.

Second, we replace the activation functions in the standard adapter with rational. The results are reported in \emph{Rational-only} rows. The results of \emph{Baseline} compared to \emph{Rational-only} show that the impact of rational is prominent when the model contains fewer parameters and using rational with an overparameterized model is not very effective, i.e., both layer selection and learnable activation play an important role.  

Third, we only add a standard adapter layer at the last 13 layers of BERT-large (\emph{Baseline$^{13}$}), which is the same number of adapter layers in \emph{AA-focused$^{sim}$}. The difference is the activation function that is used in these 13 adapter layers is ReLU in \emph{Baseline$^{13}$} and rational in \emph{AA-focused$^{sim}$}. The considerably higher performances of \emph{AA-focused$^{sim}$} show that higher performances of \emph{AA-focused} are due to both layer selection as well as a learnable activation function.

\paragraph{Learned rational activation functions.}
Figure~\ref{fig:gs} shows the learned activation functions across different layers of the same trained adapter and different tasks.
We see that the learned activation differs for different layers of the same task as well as for different tasks.

\section{Conclusion}
In this paper, we propose adaptable adapters. They consist of a learnable switch to select a subset of beneficial adapter layers and a learnable activation function to learn the suitable activation at each adapter layer and for each input data. 
The results of adaptable adapters show that we can achieve on-par performances with the full adapter architecture by using a smaller subset of layers.
We show that adaptable adapters are viable tools for designing efficient and effective adapter architectures that require less storage space, lower training and inference time with high performances.

\section*{Acknowledgements}
The authors would like to thank Jorge Cardona for his valuable contribution to the implementation of adaptable adapters. We thank Michael Bugert, Ji-Ung Lee, and Soumya Sarkar for their constructive suggestions and feedback on this work. We would like to thank Jonas Pfeiffer and Clifton Poth for always being very helpful with all questions about adapters and AdapterHub. 
This research work has been funded by the German Federal Ministry of Education and Research and the Hessian Ministry of Higher Education, Research, Science and the Arts (HMWK) within their joint support of the National Research Center for Applied Cybersecurity ATHENE. It benefited from the HMWK  cluster project ``The
Third Wave of AI''.

\bibliography{ms}
\bibliographystyle{acl_natbib}

\appendix

\section{BERT-base Results}
\insertoverallResults

\end{document}